\documentclass[journal]{IEEEtran}
\ifCLASSINFOpdf
\else
\fi
\usepackage{dblfloatfix}
\usepackage{amssymb}
\usepackage[bb=boondox]{mathalfa}
\usepackage{multirow}
\usepackage{array}
\makeatletter
\newcommand*{\rom}[1]{\expandafter\@slowromancap\romannumeral #1@}
\makeatother
\usepackage{amsmath,graphicx,subfigure,epsfig,multirow,epstopdf}
\usepackage{cite}
\usepackage{ragged2e}
\usepackage{footnote}
\usepackage{threeparttable}

\makeatletter
\newcommand{\thickhline}{
	\noalign {\ifnum 0=`}\fi \hrule height 1pt
	\futurelet \reserved@a \@xhline
}

\begin{document}
%
\title{Co-salient Object Detection Based on Deep Saliency Networks and Seed Propagation over an Integrated Graph}
%
%
%

\author{Dong-ju~Jeong,
	Insung~Hwang,
	and~Nam~Ik~Cho,~\IEEEmembership{Senior~Member,~IEEE}
\thanks{D. Jeong, I. Hwang, and N. I. Cho are with the Dept. of Electrical and Computer Engineering,
Seoul National University, 1, Gwanak-ro, Gwanak-Gu, Seoul 151-
742, Korea and also affiliated with INMC (e-mail: jeongdj@ispl.snu.ac.kr, coee55@gmail.com, and
nicho@snu.ac.kr).}}

%
%

\markboth{Journal of \LaTeX\ Class Files,~Vol.~14, No.~8, August~2015}%
{Shell \MakeLowercase{\textit{et al.}}: Bare Demo of IEEEtran.cls for IEEE Journals}
%



\maketitle

\begin{abstract}
This paper presents a co-salient object detection method to find common salient regions in a set of images. We utilize deep saliency networks to transfer co-saliency prior knowledge and better capture high-level semantic information, and the resulting initial co-saliency maps are enhanced by seed propagation steps over an integrated graph. The deep saliency networks are trained in a supervised manner to avoid online weakly supervised learning and exploit them not only to extract high-level features but also to produce both intra- and inter-image saliency maps. Through a refinement step, the initial co-saliency maps can uniformly highlight co-salient regions and locate accurate object boundaries. To handle input image groups inconsistent in size, we propose to pool multi-regional descriptors including both within-segment and within-group information. In addition, the integrated multilayer graph is constructed to find the regions that the previous steps may not detect by seed propagation with low-level descriptors. In this work, we utilize the useful complementary components of high-, low-level information, and several learning-based steps. Our experiments have demonstrated that the proposed approach outperforms comparable co-saliency detection methods on widely used public databases and can also be directly applied to co-segmentation tasks.
\end{abstract}


%
\IEEEpeerreviewmaketitle

\section{Introduction}
\IEEEPARstart{T}{he} objective of saliency detection is to find the most informative and attention-drawing regions in an image \cite{yang2013_GMR}, and it has been one of the most popular computer vision tasks for the past few decades \cite{zhang2016ldw}. There may be two categories of saliency detection: salient object detection and eye fixation prediction. The former aims at identifying precise salient object regions with relative saliency values \cite{yang2013_GMR,wang2016_Grab,liu2016_DHSNet,kuen2016_RANSD}, while the latter is for estimating eye gaze fixation points resulting in saliency maps in the form of heat maps \cite{pan2016sdc,jetley2016ete,cholakkal2016backtrack,kruthiventi2016su}. Recently, co-saliency detection has emerged as an important subtopic of the salient object detection, which is to find visually distinct regions and/or objects that commonly appear in a set of images. In other words, the goal of co-saliency detection is to find \textit{common salient} objects while suppressing salient objects/regions that appear only in part of the image group. Thus, it is needed to consider visual coherency among the images besides the cues used in the saliency detection such as contrast \cite{cheng2011_HC_RC,cheng2015_HC_RC,zhu2014saliency} and/or boundary priors \cite{zhu2014saliency,wei2012_GS,yang2013_GMR}. The co-saliency detection can be applied to other computer vision tasks, such as co-segmentation \cite{zhang2016ldw}, video foreground detection \cite{fu2015mfvcs}, image retrieval \cite{fu2013cb}, and weakly supervised localization \cite{zhang2016review}. It can be utilized to enhance the single-image saliency detection as well \cite{cheng2014salientshape}.

Many researchers have recently proposed to utilize convolutional neural networks (CNNs), which will be called deep saliency networks in this paper, to produce pixel- or segment-level saliency maps better capturing high-level semantic information and robust to complex background \cite{li2015_VSMDF,liu2016_DHSNet,kuen2016_RANSD,li2016_DCL}. These methods also detect salient regions more uniformly, and outperform conventional algorithms in terms of accuracy. Meanwhile, until recently, the majority of co-saliency detection methods use low-level handcrafted features such as color cues because the color information usually play an important role in distinguishing between co-salient and non-co-salient regions \cite{fu2013cb,liu2014hs,li2015mg,cao2014fusion}. However, recent advances in deep learning have also contributed to the state-of-the-art methods for co-saliency detection \cite{zhang2016ldw,zhang2016mil,zhang2016transfer}, which exploit high-level CNN features to represent image patches/segments or encode low-level features with deep autoencoders. One of the most challenging issues in co-saliency detection is its dependency on input image groups: whether low- or high-level features become a prior factor differs from case to case, and the same goes for contrast and consistency \cite{zhang2016review}. To handle this, those learning-based methods perform weakly supervised learning given an image group, where similar images from external groups are also exploited to identify consistent background. On the other hand, the graph-based processing proves to be effective for spatial refinement of each image \cite{zhang2016mil,li2016_DCL}, but it has rarely been used considering a whole image group and its consistency factor.

\begin{figure*}[htb!]
	\centering
	\includegraphics[width=\linewidth]{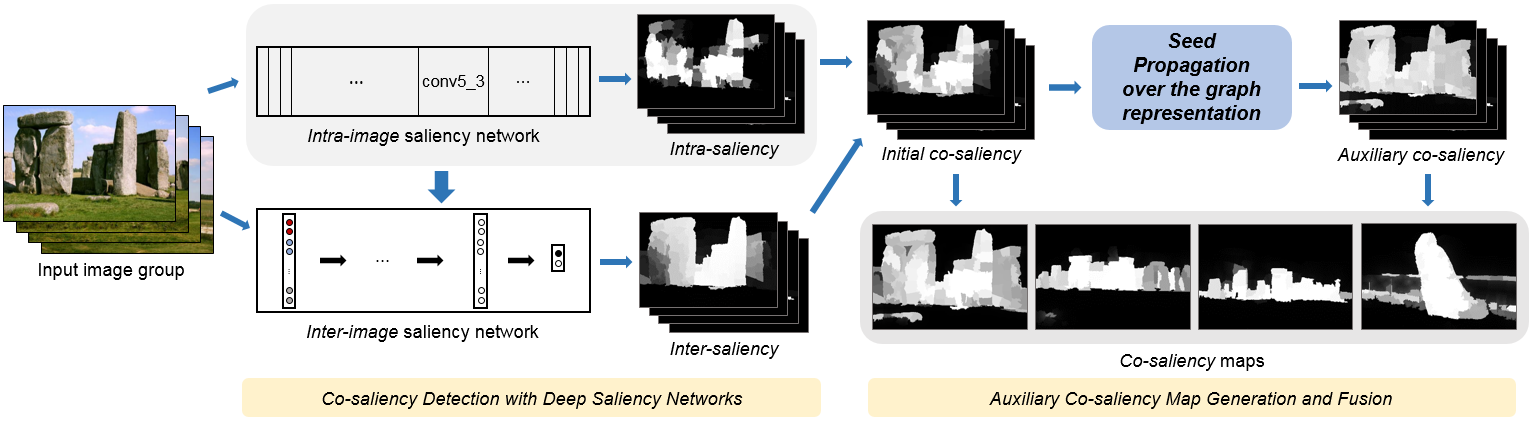}
	\caption{A flowchart of the proposed co-saliency detection method with the blocks showing its steps and produced items.}
	\label{fig:framework}
\end{figure*}

In order to tackle the issues mentioned above, we propose a supervised learning-based method that is complemented by graph-based manifold ranking with an integrated graph including all the intra-image nodes of input images. The intra-image saliency (IrIS) maps of the images are produced by a fully convolutional networks, the part of which generates high-level semantic features. They are associated with low-level features to cope with the various cases and improve the performance of our system \cite{Lee2016_ELL}, and fed into fully-connected layers to obtain the inter-image saliency (IeIS) value of each segment. We choose to train these deep saliency networks in a supervised manner to avoid using any learning models trained given an input image group (with similar external images) and thereby reduce computation time. As a result, initial co-saliency maps are generated by combining the IrIS and IeIS maps. In addition, we propose to construct the integrated graph where auxiliary co-saliency values are obtained by propagating seeds extracted from the initial co-saliency maps. The segments of the input images are treated as the intra-image nodes, and inter-image nodes connect them to form the integrated graph with a sparse affinity matrix computed with color similarities. While the deep saliency networks are expected to detect precise (co-)salient regions, the graph-based method helps to find parts of co-salient objects showing color consistency and/or located on image boundaries. These two types of co-saliency value are combined to produce final co-saliency maps with simple spatial refinement. The unified framework is illustrated in Fig.~\ref{fig:framework}.

The rest of this paper is organized as follows. The second section introduces related works on co-saliency detection, the third section describes co-saliency detection using the deep saliency networks, and the fourth section describes the seed propagation over the integrated graph. The experimental results and the conclusions are presented in the last two sections.

\section{Related Work}
The co-salient object detection began with analyzing multi-image information and finding common objects within image pairs \cite{li2011co,tan2013image,jacobs2010cosal,chen2010preatten}. For example, Li et al. \cite{li2011co} performed the pyramid decomposition of images and then extracted color and texture features from each region to compute the maximum SimRank scores of region pairs, which are defined as multi-image saliency values. To obtain the final co-saliency maps, they linearly combined the single- and multi-image saliency maps. Tan et al. \cite{tan2013image} proposed to calculate the affinities of superpixel pairs with color and position similarities, and then perform bipartite graph matching to discover the most relevant pairs for affinity propagation. The resulting superpixel affinities between two images are converted into foreground cohesiveness and locality compactness measures to obtain the final co-saliency maps.

Due to the lack of scalability, other co-saliency detection methods have aimed at treating larger groups with more than two images. Fu et al. \cite{fu2013cb} proposed a two-layer cluster-based approach, where pixel-level intra- and inter-image clustering steps are performed to calculate the contrast, spatial, and corresponding cues of each cluster. They employed multiplication fusion of the cluster-wise cues and converted them into the final pixel-wise co-saliency values. The algorithm by Li et al. \cite{li2013tmm} generates intra-saliency maps with multi-scale segmentation and pixel-wise voting, and inter-saliency maps by matching image regions with a minimum spanning tree. It also linearly combines the intra- and inter-saliency maps into the final co-saliency maps. Liu et al. \cite{liu2014hs} proposed to perform hierarchical segmentation and compute intra-saliency, object prior, and global similarity values of the fine/coarse segments to obtain co-saliency values. In \cite{li2015mg}, Li et al. adopted their previous work to obtain single-image saliency maps, and used the two-stage manifold ranking method to estimate co-salient regions. They let each image of a group take turns to produce queries for the manifold ranking of all the images, and fused multiple co-saliency values by averaging or multiplication. In addition, Cao et al. \cite{cao2014fusion} proposed a fusion-based algorithm, which adopts several existing (co-)saliency detection schemes and combine their results with self-adaptive weights produced by low-rank analysis.

The above methods utilize handcrafted features to represent pixels, segments, or clusters, and some of them focus only on color cues to cope with the situation where co-salient objects are quite consistent in color; so they cannot capture abstract semantic information and effectively detect the co-salient objects that consist of multiple components. Thus the learning-based methods using high-level features \cite{zhang2016ldw,zhang2016mil,zhang2016transfer} have recently been proposed to tackle this problem. Zhang et al. \cite{zhang2016ldw} proposed to find several similar neighbors from external groups for negative image patches and analyze intra-image contrast, intra-group consistency, and inter-group separability measures. They combined them through a Bayesian framework to obtain patch-wise co-saliency values and then converted them into pixel-wise ones. In \cite{zhang2016mil}, a self-paced multi-instance learning method is used to update positive and negative training samples and their weights, and thereby train an SVM model for co-saliency estimation, where similar neighbors give the negative samples as in \cite{zhang2016ldw}. The approach proposed in \cite{zhang2016transfer} exploits stacked denoising autoencoders (SDAEs) for two objectives: intra-saliency prior transfer and deep inter-saliency mining. First, several SDAEs for intra-saliency detection are trained in supervised and unsupervised manners with single-image saliency detection data and second, another SDAE is trained in an unsupervised manner with input images to exploit its reconstruction errors for co-saliency cues. These three approaches also utilize CNN models or SDAEs to represent patches/superpixels with high-level features or convert low-level descriptors into higher-level ones. In addition, they perform their learning steps provided with input image groups because the criteria for differentiating between co-salient and non-co-salient objects depend on the given target image group; but, this may result in high computational complexity for testing.

\section{Co-saliency detection using deep saliency networks}
According to \cite{zhang2016review}, the co-saliency detection methods in the literature explicitly or implicitly use the contrast cue and corresponding cue, which are also called intra- and inter-image saliency respectively. This is because co-salient regions are salient in each image and have correspondence in a whole image group, and thus a co-saliency detection algorithm should not ignore either one. To be accurate, the definition of the inter-image saliency in the conventional methods is similar to that of co-saliency but places more emphasis on the correspondence factor. As the bottom-up methods for co-saliency detection generally design the explicit intra- and inter-image saliency maps, we compute them with deep saliency networks trained in a supervised manner. Then, they are refined and combined to produce initial co-saliency maps for the next step.

\subsection{Intra-Image Saliency Detection}
Given an image group $\{I^m\}_{m=1}^M$, each image $I^m$ is independently represented by its $n_m$ superpixels $\{s^m_i\}_{i=1}^{n_m}$, which are over-segmented regions obtained by the SLIC algorithm \cite{achanta2012_SLIC}. The goal of this step is to produce pixel-wise intra-image saliency (IrIS) values and convert them into segment-wise ones $\{rs^m_i\}_{i=1}^{n_m}$ for each image $I^m$. To this end, we use the multi-scale fully convolutional network \cite{li2016_DCL}, which produces pixel-level saliency maps combining several stacked feature maps extracted with different sizes of receptive fields. We call this single-image saliency detection network as an IrISnet in this paper. It is based on the original structure that utilizes the pre-trained VGG16 network \cite{simonyan2015vgg} and is implemented following the DeepLab system \cite{chen2017deeplab}.

\begin{figure}[htb!]
	\centering
	\includegraphics[width=0.24\linewidth]{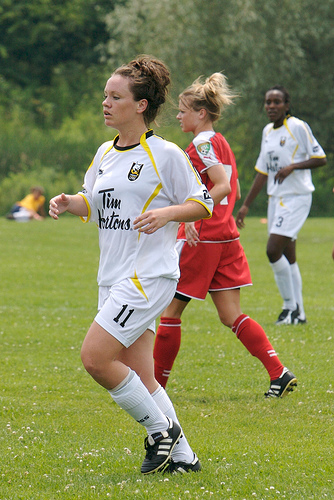}
	\includegraphics[width=0.24\linewidth]{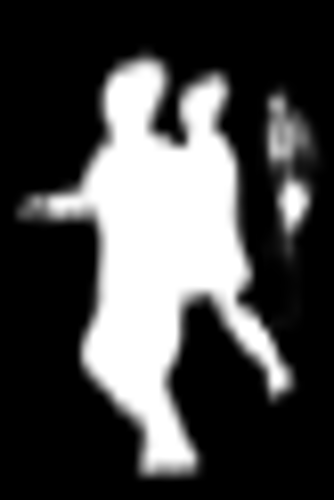}	\includegraphics[width=0.24\linewidth]{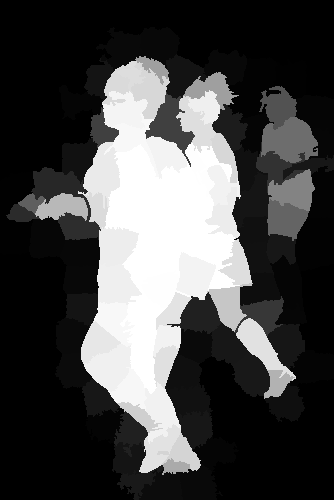}
	\includegraphics[width=0.24\linewidth]{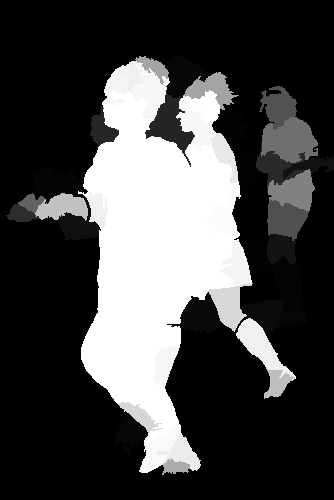}
	\caption{Examples of the intra-image saliency maps. From left to right: an input image, its pixel-level IrIS map, and two segment-level IrIS maps. The third and fourth images show the mean and median of the pixel-wise saliency values within each segment, respectively.}
	\label{fig:IrIS}
\end{figure}

Specifically, it replaces the fully-connected layers of the original VGG16 network with $1\times1$ convolutional layers to design its fully convolutional structure, and four branches consisting of $3\times3$ and $1\times1$ convolutional layers are attached to its pooling layers to obtain the multi-scale feature maps. The main stream and four branches compute the five multi-scale single-channel feature maps, which are input to the last $1\times1$ convolutional layer and a sigmoid activation function to obtain an output saliency map ranging between $0$-$1$. This network also exploits the hole (\`a trous) algorithm \cite{holschneider1990hole} for two purposes: first it helps to compute denser feature maps maintaining the original sizes of receptive fields and second, it can also adjust the size of each multi-scale feature map to be identical. As a result, we obtain the output maps for $\{I^m\}$ and use bicubic interpolation to resize them to the original input image sizes so that we can estimate pixel-level saliency maps. Lastly, we set the median of the saliency values within each superpixel as $rs_i^m$ to obtain the segment-wise IrIS values, where we use the medians instead of means to reduce halos around salient objects as shown in Fig.~\ref{fig:IrIS}.

\subsection{Inter-Image Saliency Detection}
For both single-image salient object detection and co-salient object detection, many of the existing methods (over-)segment input images and produce segment-wise saliency values. As for graph-based models \cite{yang2013_GMR, wang2016_Grab}, superpixels instead of raw pixels are treated as nodes in a graph, the number of which is limited, so this makes it possible to utilize the graph models by manifold ranking. Meanwhile, the deep convolutional networks can make it possible to produce pixel-level saliency maps, and some CNN-based methods efficiently perform in that manner \cite{liu2016_DHSNet,kuen2016_RANSD}. However, other ones operate totally at segment-level or use segment-wise saliency values to complement pixel-wise ones. In \cite{li2015_VSMDF,Lee2016_ELL}, each segment (with its relevant regions), irrespective of the number of segments, can be fed into the deep neural networks with the fixed number of parameters, and low-level features can also be exploited as additional inputs. The method of \cite{li2016_DCL} utilizes both the pixel- and segment-wise saliency values, where the latter ones better represent saliency discontinuities along object boundaries.

To treat multiple images, we take advantages of the segment-wise processing mentioned above. When a CNN-based method is applied, as for single-image saliency detection, a whole image can be fed into a CNN model. However, for co-saliency detection, the size of an image group is not consistent and the information of a whole image group has to be exploited, so it is appropriate to predict segment-wise co-saliency values. In addition, there are cases when color cues are the most important rather than the other ones such as high-level semantic information, and other low-level features (e.g., position) might also be helpful and need to be added to the higher-level features extracted from convolutional layers. Considering these aspects, we compute each segment's descriptor that includes the information of a whole image group, and produce segment-level inter-image saliency (IeIS) maps.

The CNN features extracted from the conv5\_3 layer within the IrISnet are used as higher-level features for each segment. To adapt the superpixels made on the image domain to the domain of the feature maps of the IrISnet, we use convolutional feature masking (CFM) \cite{dai2015_CFM}, and then perform $2\times2$ spatial pooling \cite{he2015spp} to obtain fixed-length descriptors as in \cite{li2016_DCL}. In addition to the higher-level features, the low-level ones such as \textit{Lab} color vectors, color histograms, and positions are computed to complement the higher-level features because either one of the two types is more important than the other one depending on the input image group \cite{zhang2016review}. With these components, we propose to compute each segment $s^m_i$'s multi-regional descriptor $\mathbf{x}^m_i = [\mathbf{x}^m_{i,\text{seg}}, \mathbf{x}^m_{i,\text{nbh}}, \mathbf{x}^m_{\text{sfg}}, \mathbf{x}_{\text{gfg}}]$, each element of which is pooled within four different regions: i) the target segment $s_i^m$, ii) its immediate neighborhood, iii) foreground regions in the image that it belongs to, and iv) foreground regions in the whole image group. Each image $I^m$ can initially have one or multiple foreground regions, which are set by thresholding its IrIS map with $\text{max}(\frac{1}{n_m}\sum_{i=1}^{n_m}rs_i^m,0.5)$ and finding connected components. Then we form the power set of them, where the empty set is excluded, and all its elements are treated as the foreground regions of the image. The \textit{Lab} color vectors and positions are normalized to $[0,1]$ and averaged to represent each region. Also, each foreground region has the variance of the $(x,y)$-positions within the region. We $L1$-normalize and then square root the 256-bin \textit{Lab} color histograms as RootSIFT \cite{arandjelovic2012rootsift} and VLAD \cite{jegou2012vlad} to moderately suppress the few  color components bursty in the image group. Given the descriptors of all the foreground regions, we perform sum-pooling to obtain fixed-length $\mathbf{x}^m_{\text{sfg}}$ and $\mathbf{x}_{\text{gfg}}$. In particular, the sum-pooling of regional max-pooled CNN features has been shown to be effective in \cite{gordo2016deepretrieval,tolias2016rmac}, but the difference from R-MAC \cite{tolias2016rmac} is that we perform the $2\times2$ spatial pooling over the fixed grid in each region. We compute the covariance matrices of the high- and low-level descriptors within the foreground regions, and the traces of them are included in $\mathbf{x}_{\text{gfg}}$. At last, each of $\mathbf{x}^m_{i,\text{seg}}$, $\mathbf{x}^m_{i,\text{nbh}}$, $\mathbf{x}^m_{\text{sfg}}$, and $\mathbf{x}_{\text{gfg}}$ is $L2$-normalized and then they are concatenated to form $\mathbf{x}_i^m$.

\begin{figure}[htb!]
	\centering
	\includegraphics[width=\linewidth]{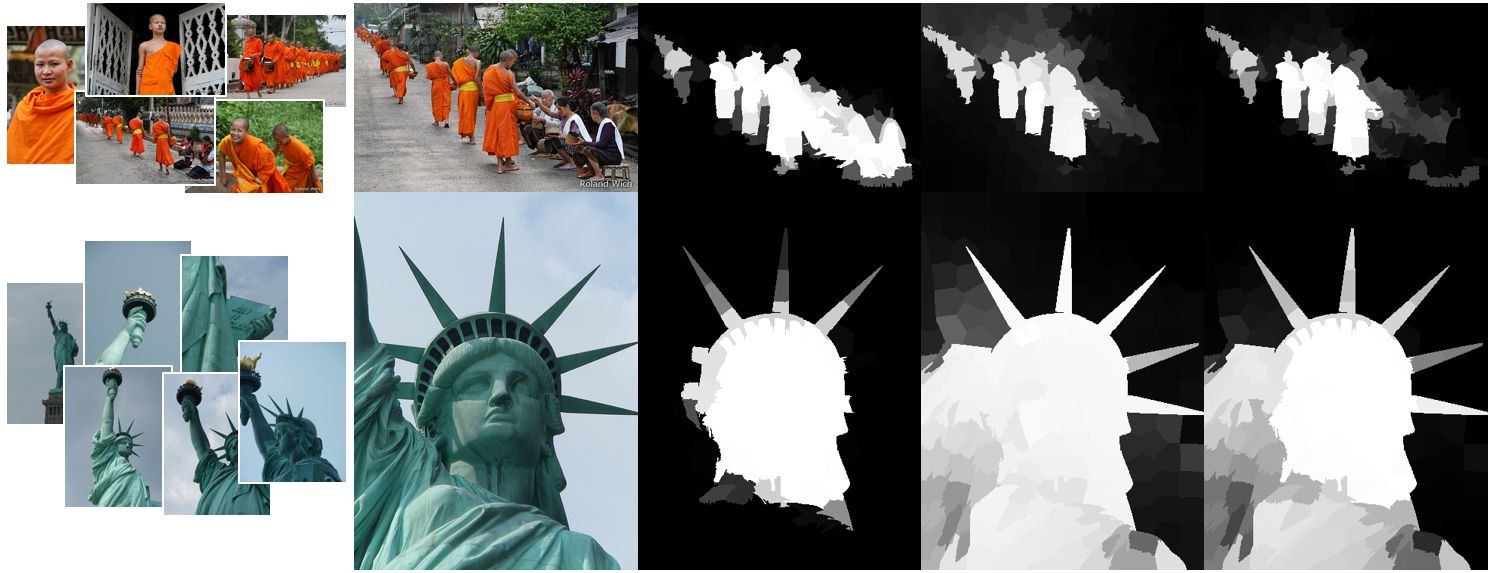}
	\caption{Examples of two different cases where IrIS and IeIS maps differ from each other. From left to right: image groups, input images from these groups, and their IrIS, IeIS, and initial co-saliency maps. The first row shows the regions that are salient in their image and not co-salient in the group, while the second row represents an opposite case.}
	\label{fig:initcosal}
\end{figure}

Given the segment descriptors $\{\mathbf{x}^m_i\}_{i=1}^{n_m}$ for each image $I^m$, they are fed into three fully-connected layers, which outputs with a two-way softmax. We call this network model an IeISnet. To train the IeISnet, the ground-truth co-saliency maps of training datasets are set to labels $cs_i^{GT}$ by thresholding the averaged label with 0.5 in each superpixel, and the cross entropy loss is used with pointwise weights as below:
\begin{align}
\begin{split}
L &= - \frac{1}{N} \sum_{i}^{N} \lambda_i \text{log} \frac{e^{z_i^{y_i}}}{e^{z_i^0}+e^{z_i^1}}, \\
\lambda_i &= \{ (1-\rho)[y_i = 0] + \rho[y_i = 1] \} \cdot \gamma^{|rs_i - cs_i^{GT}|}, \; \forall i
\end{split}
\end{align}
where $N$ is the number of training data, ${z_i^0}$ and ${z_i^1}$ are the two last activation values, $0$ and $1$ are the labels for non-co-salient and co-salient regions respectively, and $y_i$ is the ground-truth label of the $i$-th sample. The weight $\rho$ balances the number of $0$ and $1$ labels in the training sets. In terms of $\gamma$, the pointwise weights $\lambda_i$ are designed to place more emphasis on the regions that have high IrIS and low IeIS values, and vice versa. As shown in Fig.~\ref{fig:initcosal}, the former case shows what are called ``single saliency residuals" in \cite{huang2017_CFR}, and the latter one represents the situations where some regions may not seem salient in their image but are certainly co-salient in their image group.

Because the IeIS value of each segment $s^m_i$ is independently estimated through the above process, we refine each IeIS map so that neighboring regions have smooth IeIS values. For this, we perform seed propagation over a simple graph model. The segments $s^m_i$ are treated as intra-image nodes $v^m_i$ in each image $I^m$, and the edge $e_{ij}^m$ between two neighboring nodes $v_i^m$ and $v_j^m$ that share a common boundary of segments connects them with a weight $w_{ij}^m$, which represents the affinity between them and is calculated using color similarity \cite{hwang2016saliency}. Even though the recently proposed saliency detection methods using the graph-based manifold ranking \cite{yang2013_GMR, wang2016_Grab} utilize more sophisticated graphs for difficult cases such as where parts of salient objects are located on image boundaries, we tackle this problem in section \ref{sec:seedpropa} and use the simple graph model for this step. We let $\mathbf{x}_{i,co}$ denote (omitting $m$ for now) the averaged \textit{Lab} color vector of $v_i$ in an image, the weight $w_{ij}$ is computed as:
\begin{align}
\begin{split}
w_{ij} &= \exp\left( -(\mathbf{x}_{i,co}-\mathbf{x}_{j,co})^T \Sigma^{-1} (\mathbf{x}_{i,co}-\mathbf{x}_{j,co}) \right) \\
\Sigma &= \frac{1}{N(E)} \sum_{e_{ij} \in E}{(\mathbf{x}_{i,co}-\mathbf{x}_{j,co})(\mathbf{x}_{i,co}-\mathbf{x}_{j,co})^T}
\label{eq:w_ij}
\end{split}
\end{align}
where $E$ is the set of all the edges in the image and $N(E)$ is the size of $E$. Then the affinity matrix for the intra-image graph of $I^m$ is constructed whose $(i,j)$-th element is the weight between $v_i^m$ and $v_j^m$:
\begin{equation}
(\mathbf{W}^m)_{i,j}= \left\{ \begin{array}{ll}
w_{ij}^m, & \textrm{if } j \in Q_i^m,\\
0, & \textrm{otherwise},
\end{array} \right. m=1,\dots,M
\label{eq:W^m}
\end{equation}
where $Q_i^m$ is an index set of neighbors of the $i$-th node.

To propagate seeds over these graphs, we need to extract foreground and background seeds. We set the segments whose initial IeIS values are larger than 0.5 and, at the same time, in the top 10 percent as the foreground seeds, where they have $1$s and all the others have $0$s in $\mathbf{y}^m_f$. The segments on image boundaries are simply selected as the background seeds and we get $\mathbf{y}^m_b$ likewise. To obtain the refined IeIS maps, the graph-based learning method is adopted for effective propagation \cite{zhou2004learning,yang2013_GMR}. Given the weight matrix $\mathbf{W}^m$ and its degree matrix $\mathbf{D}^m = \text{diag}(d_1^m,...,d_{n_m}^m)$, where $d_i^m=\sum_{j}{w_{ij}^m}$, the newly ranked values $\mathbf{f}^m = [f_1^m,...,f_{n_m}^m]^T$ for either type of the seeds can be optimized with the following problem:
\begin{eqnarray} \label{eq:opt_GMR}
\underset{\mathbf{\mathbf{f}^m}}{\text{min}} \;
\frac{1}{2} \left( \sum_{i,j=1}^{n_m} w_{ij}^m \left| \frac{f_i^m}{\sqrt{d_i^m}} - \frac{f_j^m}{\sqrt{d_j^m}}  \right|^2 + \nu \sum_{i=1}^{n_m} \left| f_i^m - y_i^m \right|^2 \right)
\end{eqnarray}
where $\nu$ is the controlling parameter that balances the smoothness constraint and the fitting constraint. The solution of (\ref{eq:opt_GMR}) is given by:
\begin{equation}
\mathbf{f}^m = \left(\mathbf{D}^m - \alpha \mathbf{W}^m\right)^{-1} \mathbf{y}^m
= \mathbf{W}^m_L \mathbf{y}^m
\end{equation}
where $\alpha = 1/(\nu+1)$ and, for implementation, the diagonal elements of $\mathbf{W}_L^m$ are set to $0$ for each query to obtain the propagated values ranked by the other ones except the query itself. Using both the foreground and background seeds, $\mathbf{f}^m_f$ and $\mathbf{f}^m_b$ are obtained where each node receives newly propagated values from the seeds with the learned affinity matrix. The final refined IeIS values are computed as:
\begin{equation}
\mathbf{es}^m=\left(\mathbf{f}_f^m-\eta\mathbf{f}_b^m\right)./ \left(\mathbf{f}_f^m+\eta\mathbf{f}_b^m\right) = [es_i^m,...,es_{n_m}^m]^T
\end{equation}
where $./$ is the element-wise division of two vectors and $\eta$ is a controlling parameter. The numerator represents the IeIS while the denominator maintains the balance among the nodes, and lastly $\mathbf{es}^m$ is normalized to $[0,1]$.

\subsection{Initial Co-saliency Maps}
As mentioned above, there are the occasions where $rs_i^m$ should be sufficiently larger than $es_i^m$, and vice versa. If $rs_i^m > es_i^m$, $s_i^m$ is considered to show the single saliency residual and thus the co-saliency value of $s_i^m$ should be as small as $es_i^m$, which prohibits us from linear combination of the two values \cite{huang2017_CFR}. If $es_i^m > rs_i^m$, on the other hand, this shows the specific case where some regions may not seem salient in their image but are certainly co-salient in their image group. Both the types of cases encourage us to put more emphasis on $es_i^m$ for computing co-saliency maps. Considering this aspect, we obtain the initial co-saliency (IC) value for each segment $s_i^m$ as below:
\begin{align}
\begin{split}
IC_i^m &= \left\{ \begin{array}{ll}
rs_i^m \cdot es_i^m, & \delta_i^m \ge \tau \\
\left(1-|\delta_i^m|\right)rs_i^m + |\delta_i^m|es_i^m, & \text{otherwise}
\end{array} \right. \\
\delta_i^m &= rs_i^m - es_i^m.
\label{eq:initcosal}
\end{split}
\end{align}
where the threshold $\tau$ draws a boundary between the ``single saliency residual" case and the other one. Fig.~\ref{fig:initcosal} shows several saliency maps resulted from the deep saliency networks.

\section{Seed propagation over an integrated graph}
\label{sec:seedpropa}

The (co-)saliency detection methods usually perform their latter tasks to obtain final (co-)saliency maps leveraging color and pixel position information. For example, the ranking with foreground queries in \cite{yang2013_GMR} is the second stage to locate accurate object boundaries and eliminate background noise, and a fully-connected conditional random field (CRF) \cite{krahenbuhl2011_CRF} is used for post-processing in \cite{li2016_DCL}. Many co-saliency detection algorithms also refine their resulting maps \cite{zhang2016ldw,zhang2016mil} or combine several cues \cite{liu2014hs} using color features within an image group because co-salient objects probably share similar color features in (part of) the image group. However, the graph-based procedures among these latter tasks are performed respectively within each image, so they tackle only the refinement of each (co-)\\saliency map so that it shows accurate boundaries and has smooth saliency values, not considering the correspondence within the image group.

\begin{figure}[tb!]
	\centering
	\includegraphics[width=\linewidth]{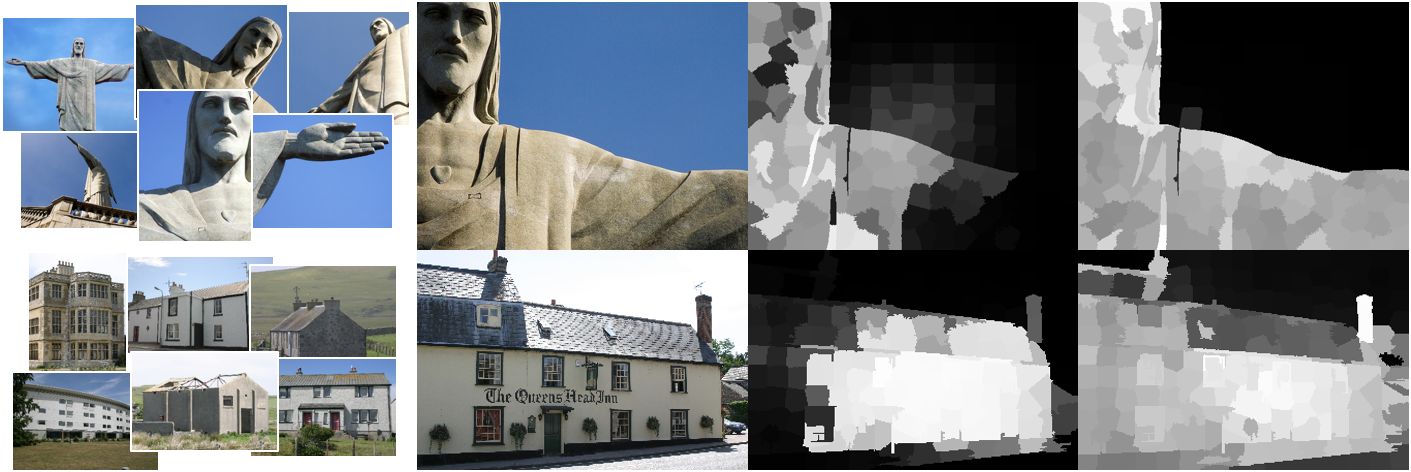}
	\caption{Examples of initial and auxiliary co-saliency maps. From left to right: image groups, input images from these groups, and their initial and auxiliary co-saliency maps. These initial co-saliency maps do not fully detect the regions that are homogeneous and/or close to image boundaries, while the second row shows that the auxiliary co-saliency maps may miss part of the objects with multiple components. Thus, the two types of co-saliency maps can complement each other.}
	\label{fig:seedpropa}
\end{figure}

In contrast to those methods, we propose to consider the whole image group and refine the input images all together. In addition, this step has another important role, which is to detect (parts of) co-salient objects located on image boundaries as shown in Fig.~\ref{fig:seedpropa}. The above procedures in our work might miss the homogeneous parts of co-salient objects on the image boundaries and strongly suppress the regions close to the boundaries. To this end, we construct an integrated graph so that it can connect all the intra-image nodes of the images in the group for sharing co-saliency information.

\subsection{The Integrated Graph with a Cluster Layer}
\begin{figure}[htb!]
	\centering
	\includegraphics[width=\linewidth]{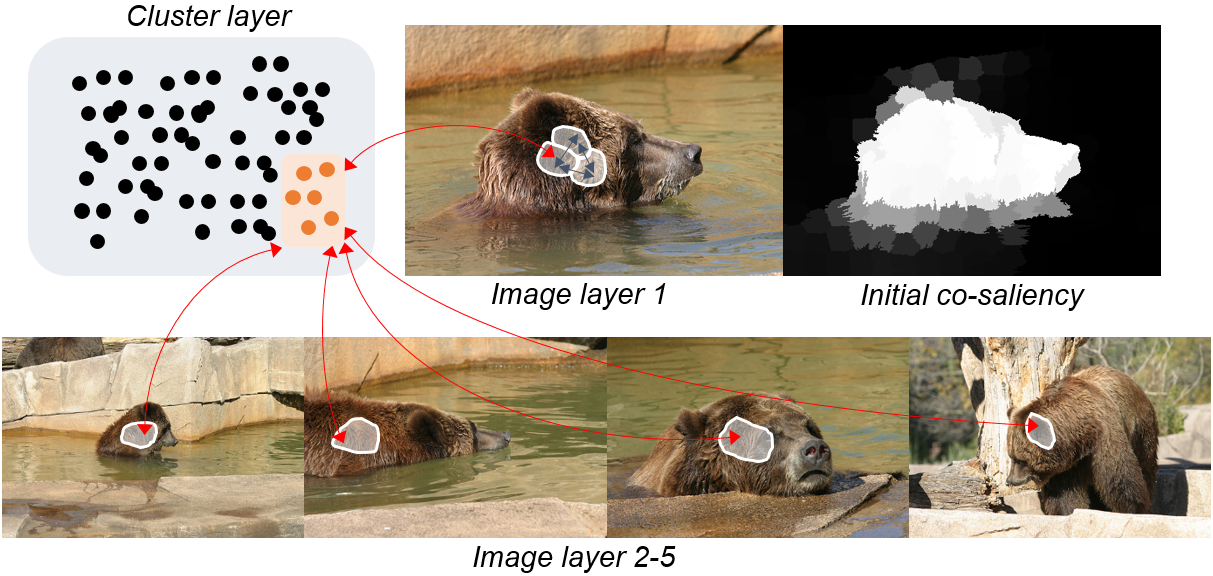}
	\caption{Visualization of the integrated graph and the interactions of intra-image nodes therein, focusing on \textit{image layer $1$}. Red arrows represent the paths where the intra-image nodes interact between images via inter-image nodes, and small navy arrows indicate the interactions between the intra-image nodes in the single image.}
	\label{fig:intgraph}
\end{figure}

In \cite{tan2013image}, the bipartite graph matching method finds pairs of the most relevant superpixels between two images, each of which is connected with its matching score. Though ensuring good matched pairs for similar scenes such as sequential frames of a video that are not severely different from each other, in general, this approach easily fails to find good pairs of superpixels between the images that have various backgrounds and/or different sizes of objects. Hence, an indirect approach is introduced in this paper to overcome this problem. We basically ignore the connectivity between images, which means that there are no edges that directly connect the intra-image nodes of any two different images. Thus the intra-image graphs are represented in the form of a sparse block-wise diagonal matrix:
\begin{equation}
\mathbf{W}_{I} = 
\begin{bmatrix}
\mathbf{W}^{1}  & 0  & 0\\
0  & \ddots & 0\\
0  & 0 & \mathbf{W}^{M}\\
\end{bmatrix} \in \mathbb{R}^{n \times n}
\label{eq:W_I}
\end{equation}
where $n=\sum_{m}n_m$ and each $\mathbf{W}^m$ is computed by (\ref{eq:w_ij},\ref{eq:W^m}). Instead, the proposed method introduces an additional \textit{cluster layer} to consider the interactions between images and indirectly connect the intra-image nodes via the inter-image ones on it, as shown in Fig.~\ref{fig:intgraph}.

To define the inter-image nodes, we perform $K$-means clustering with the descriptor of every intra-image node, reusing its averaged \textit{Lab} color vector $\mathbf{x}_{i,co}$. Through this procedure, $K$ clusters $\{\mathbf{C}_i\}_{i=1}^K$ and their centroids $\{\mathbf{c}_i\}_{i=1}^K$ are generated, where $\mathbf{c}_i$ is the representative descriptor for $\mathbf{C}_i$ and also defined as an inter-image node. The goal of this step is to construct the affinity matrix of the unified graph including all the intra- and inter-image nodes, so we first connect each $\mathbf{c}_i$ to its elements and compute the weights of the edges using descriptor similarities as:
\begin{align}
\begin{split}
w_{ij}^{IC} &= \exp \left(-\frac{\left\|\mathbf{x}_{i,co} - \mathbf{c}_j\right\|_2}{\sigma} \right) \\
(\mathbf{W}_{IC})_{i,j} &= \left\{ \begin{array}{ll}
w_{ij}^{IC}, & \textrm{if } \mathbf{x}_{i,co} \in \mathbf{C}_j \\
0, & \textrm{otherwise}
\end{array} \right.
\end{split}
\end{align}
where $\sigma$ is a control parameter for the descriptor similarity. In addition, the inter-image nodes are also connected to each other, specifically to their $k$-nearest neighbors ($k$-NN), which means that the graph of the cluster layer is as sparse as the intra-image graphs, and its affinity matrix is written as:
\begin{align}
\begin{split}
w_{ij}^{C} &= \exp \left(-\frac{\left\|\mathbf{c}_i - \mathbf{c}_j\right\|_2}{\sigma} \right) \\
\mathbf{W}_{C} &= \left\{ \begin{array}{ll}
w_{ij}^C, & \textrm{if } i \in k\textrm{-NN}(j) \textrm{  or  } j \in k\textrm{-NN}(i) \\
0, & \textrm{otherwise.}
\end{array} \right.
\end{split}
\end{align}
Finally, the affinity matrix of the unified graph is constructed from $\mathbf{W}_I$, $\mathbf{W}_{IC}$, and $\mathbf{W}_C$, expressed in a block-wise matrix form:
\begin{equation}
\mathbf{W} = 
\begin{bmatrix}
\mathbf{W}_{I}  & \mathbf{W}_{IC}\\
\mathbf{W}_{IC}^T  & \mathbf{W}_{C}\\
\end{bmatrix} \in \mathbb{R}^{(n+K)\times(n+K)}.
\label{eq:W_f}
\end{equation}

\subsection{Seed Propagation}
To assign newly propagated co-saliency values to all the segments, we need to extract foreground seeds (called co-saliency seeds in this section) and background ones, which are selected similarly to the process for the IeIS refinement. The top 10 percent of co-salient regions with respect to the IC values in each image are extracted as the co-saliency seeds, and the boundary nodes of each image are selected as the background seeds, based on the boundary prior. In addition, the ones selected as both the co-saliency and background seeds simultaneously are precluded from both seed sets because those seeds are not reliable. In summary, the co-saliency and background seeds are defined as:
\begin{itemize}
	\item Co-saliency seeds ($\mathbf{y}_{I,s}$) : high IC nodes that are not on any image boundaries.
	\item Background seeds ($\mathbf{y}_{I,b}$) : low IC nodes on image boundaries.
\end{itemize}

From the co-saliency and background seeds, co-saliency values are computed by propagating them to all the (intra-image) nodes in the image group. For this, we use the graph-based learning scheme again with the integrated graph, which makes a full pairwise graph as:
\begin{equation}
\mathbf{W}_{L} =\left(\mathbf{D}-\alpha \mathbf{W}\right)^{-1}
=\left[\mathbf{w}_L^1,...,\mathbf{w}_L^{n+K}\right],
\label{eq:W_LL}
\end{equation}
where $\mathbf{D}=\text{diag}(d_{1},...,d_{n})$ is the degree matrix of $\mathbf{W}$. As mentioned above, there are no direct inter-image connections between any two intra-image nodes in the graph with the affinity matrix $\mathbf{W}$, so the inter-image nodes indirectly connects the pairs of them instead. However, the learned graph with $\mathbf{W}_L$ has full pairwise relations of all the nodes. In other words, this graph has direct inter-image connections so that it ensures straightforward propagation between images.

To obtain auxiliary co-saliency maps to combine with the IC maps, the overall affinities to the co-saliency and background seeds are computed respectively, which is written as:
\begin{equation}
\mathbf{f}_{s}=\mathbf{W}_{L} ~ \mathbf{y}_s=\sum_{i\in \mathbf{S}_s}{\mathbf{w}_{L}^{i}}, \; 
\mathbf{f}_{b}=\mathbf{W}_{L} ~ \mathbf{y}_b=\sum_{i\in \mathbf{S}_b}{\mathbf{w}_{L}^{i}}
\label{eq:fsfb_cosaliency}
\end{equation}
where $\mathbf{y}_s=\left[\mathbf{y}_{I,s}; \mathbf{0}\right]$ and $\mathbf{y}_b=\left[\mathbf{y}_{I,b}; \mathbf{0}\right]$ are the co-saliency and background seed vectors respectively each of which is concatenated with a zero vector for the inter-image nodes, and $\mathbf{S}_s$ and $\mathbf{S}_b$ represent the co-saliency and background seed sets respectively. $\mathbf{f}_s$ and $\mathbf{f}_b$ are decomposed into the vectors for each image and the cluster layer, i.e., $\mathbf{f}_s=\left[\mathbf{f}_s^{1};...;\mathbf{f}_s^{M}; \mathbf{f}_s^{C}\right]$ and $\mathbf{f}_b=\left[\mathbf{f}_b^{1};...;\mathbf{f}_b^{M}; \mathbf{f}_b^{C}\right]$, and thus the auxiliary co-saliency map for $I^m$ is computed as:
\begin{equation}
\mathbf{AC}^m=\left(\mathbf{f}_s^m-\eta\mathbf{f}_b^m\right)./ \left(\mathbf{f}_s^m+\eta\mathbf{f}_b^m\right).
\label{eq:co-saliency}
\end{equation}
Lastly, $\mathbf{AC}^m = [AC_1^m,...,AC_{n_m}^m]$ is also normalized to $[0,1]$ and combined with $\mathbf{IC}^m = [IC_1^m,...,IC_{n_m}^m]$.

\subsection{Final Co-saliency Maps}
Given the initial and auxiliary co-saliency maps, the former ones might not fully detect the regions that are homogeneous and/or close to image boundaries, while the latter ones might miss part of the objects with multiple components due to solely using the color and position cues. Therefore, these are complementary to each other and thus simply combined to produce the final co-saliency maps $\mathbf{CS}^m = [CS_1^m,...,CS_{n_m}^m]$ as
\begin{equation}
CS_i^m = \text{max} \left(IC_i^m,AC_i^m\right).
\end{equation}
Because the auxiliary co-saliency maps are likely to be vulnerable to background noise, we perform a simple post-processing scheme for $\mathbf{CS}^m$ where the outputs never exceed the inputs \cite{cao2014fusion}. This step needs spatial positional distance maps, and they can be computed with shrunk input images to reduce processing time.

\begin{figure*}[htb!]
	\centering
	
	\includegraphics[width=0.9\linewidth]{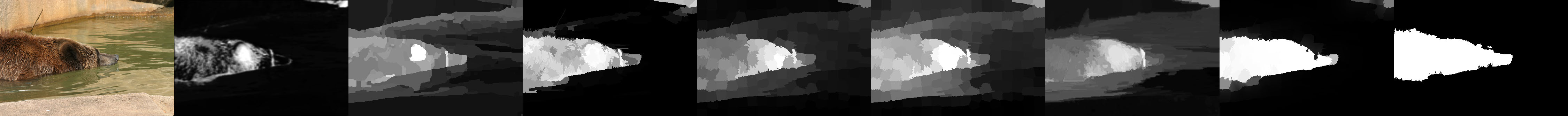} \\
	\includegraphics[width=0.9\linewidth]{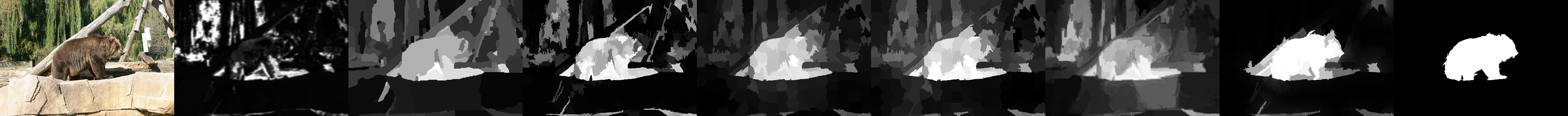} \\
	\includegraphics[width=0.9\linewidth]{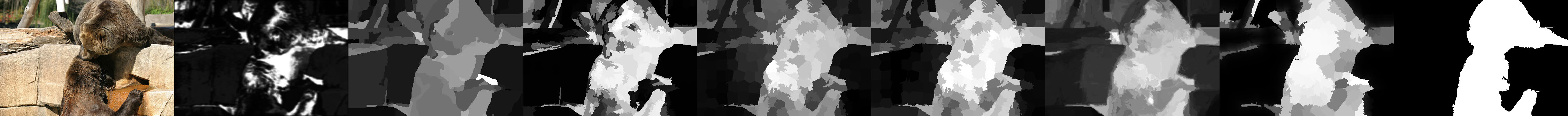} \\
	\includegraphics[width=0.9\linewidth]{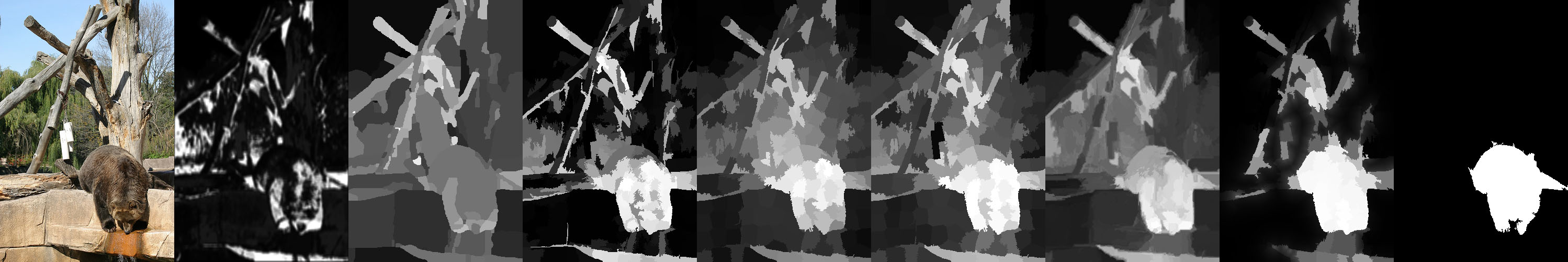} \\
	(a) \textit{Alaskan bear} \vspace{0.3em}
	
	\includegraphics[width=0.9\linewidth]{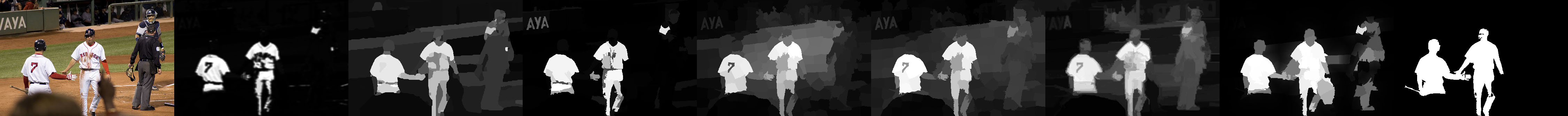} \\
	\includegraphics[width=0.9\linewidth]{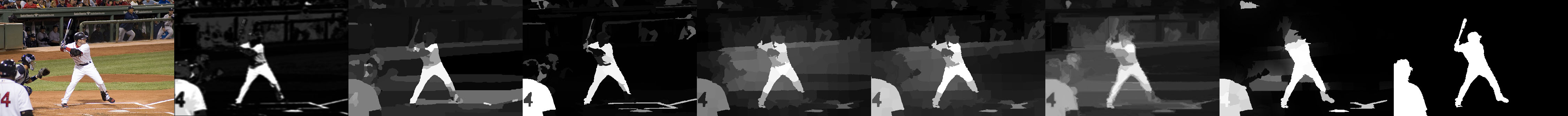} \\
	\includegraphics[width=0.9\linewidth]{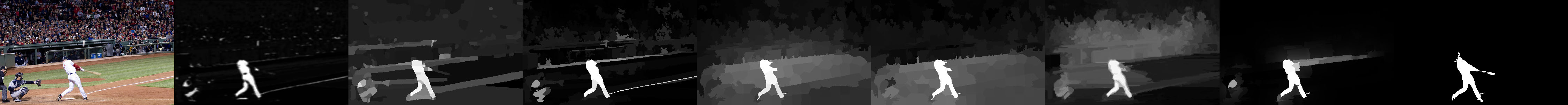} \\
	(b) \textit{Red Sox player} \vspace{0.3em}
	
	\includegraphics[width=0.9\linewidth,height=0.1\linewidth]{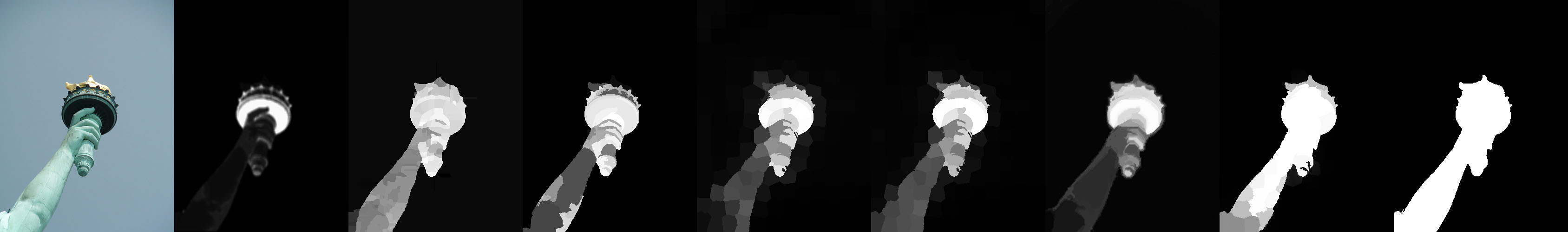} \\
	\includegraphics[width=0.9\linewidth,height=0.1\linewidth]{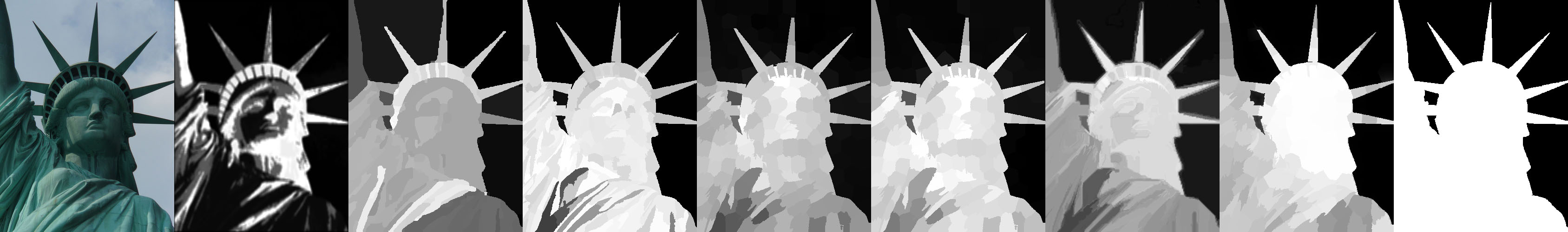} \\
	\includegraphics[width=0.9\linewidth,height=0.1\linewidth]{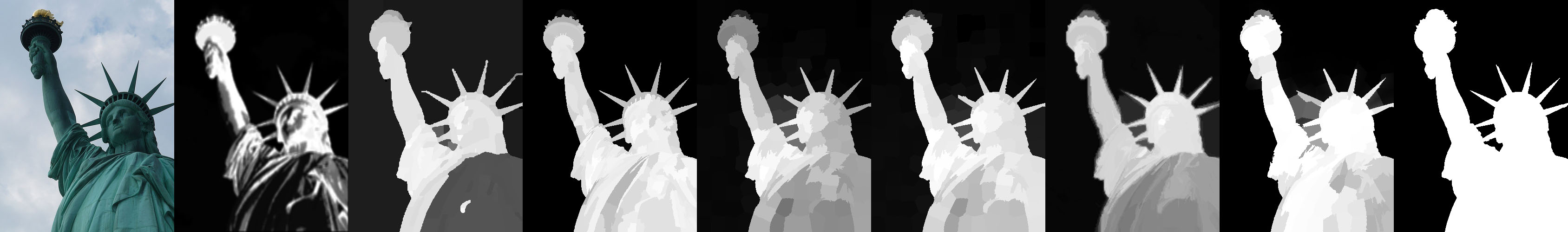} \\
	(c) \textit{Statue of liberty}
	
	\caption{Visual comparison on the \textit{Alaskan bear}, \textit{Red Sox player}, and \textit{Statue of liberty} sets in iCoseg (from left to right: input images, CB, HS, MG, LDW, MIL, DIM, the proposed method, and ground truth images).}
	\label{fig:viscmp_icoseg}
\end{figure*}

\begin{figure*}[htb!]
	\centering
	
	\includegraphics[width=0.9\linewidth]{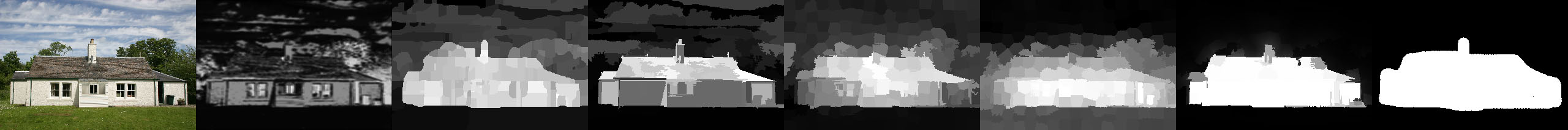} \\
	\includegraphics[width=0.9\linewidth]{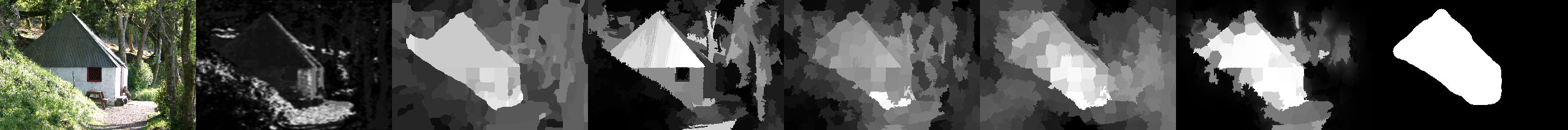} \\
	\includegraphics[width=0.9\linewidth]{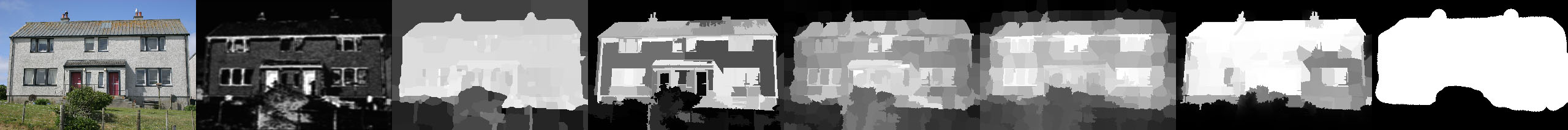} \\
	\includegraphics[width=0.9\linewidth]{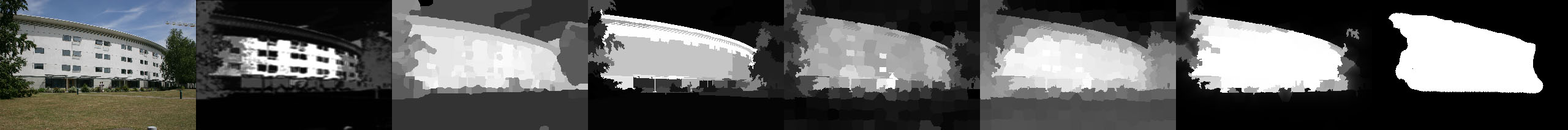} \\
	\includegraphics[width=0.9\linewidth]{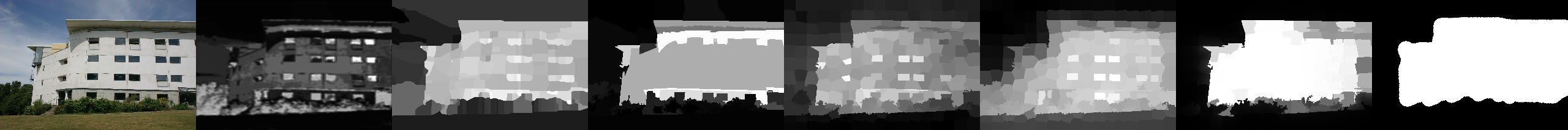} \\
	(a) \textit{Building} \vspace{0.3em}
	
	\includegraphics[width=0.9\linewidth]{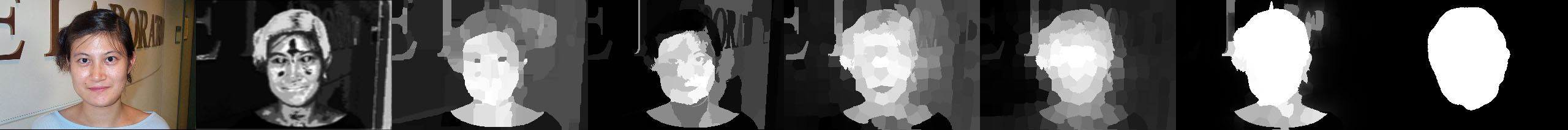} \\
	\includegraphics[width=0.9\linewidth]{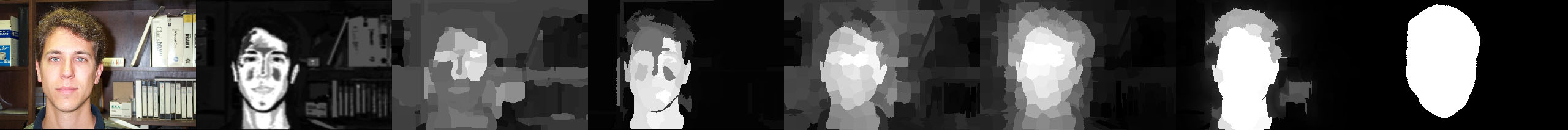} \\
	\includegraphics[width=0.9\linewidth]{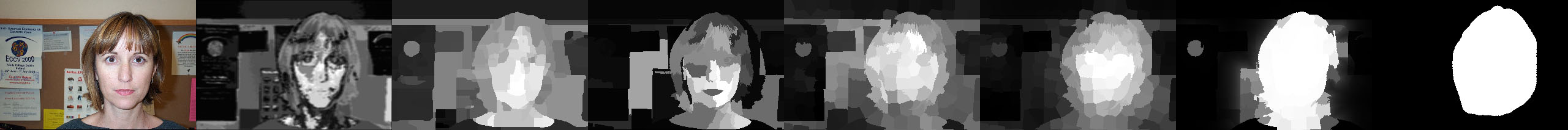} \\
	\includegraphics[width=0.9\linewidth]{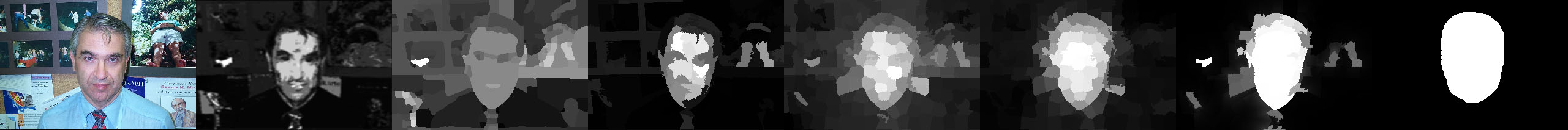} \\
	\includegraphics[width=0.9\linewidth]{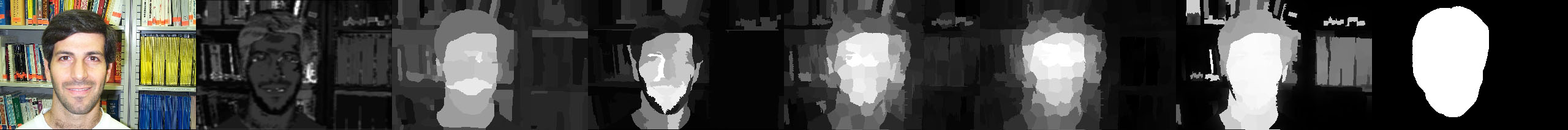} \\
	(b) \textit{Face}
	
	\caption{Visual comparison on the \textit{Building} and \textit{Face} sets in MSRC (from left to right: input images, CB, HS, MG, LDW, MIL, the proposed method, and ground truth images).}
	\label{fig:viscmp_msrc}
\end{figure*}

\section{Experimental Results}

\subsection{Experimental Settings}
In our experiments, two widely used datasets, iCoseg \cite{batra2010icoseg} and MSRC \cite{winn2005msrc}, are used to evaluate the performance of our algorithm and compare it with others. The iCoseg dataset consists of 38 groups, each of which includes 4-42 images, and totally 643 images along with pixel-wise ground truth annotations. It is the largest among widely used co-saliency detection datasets, and its image groups contain multiple objects and complex backgrounds. The MSRC dataset is composed of 8 groups, each of which equally has 30 images, but the \textit{grass} group is not used for the evaluation since it has no co-salient objects.
This dataset can be used to evaluate the ability to treat the co-salient objects that are not consistent in color, and also contains complex co-salient objects, and diverse and cluttered backgrounds.

For each of the evaluation datasets, the performance is measured with five widely used criteria: the precision-recall (PR) curve, the average precision (AP), the receiver operating characteristic (ROC) curve, areas under the ROC curve (AUC), and the F-measure. When there are more true negatives than true positives, the PR curve more clearly shows the differences between algorithms than the ROC curve does, and the same goes for their areas under the curves, AP and AUC. As for the PR and ROC curves, the co-saliency maps are normalized to $[0,255]$ and binarized with thresholds varying from 0 to 255. The precision, recall, false positive rates are calculated under each threshold and averaged over all samples as the standard used in the literature \cite{li2013evalcode}. Meanwhile, we used a self-adaptive threshold $T=\mu+\epsilon$ \cite{jia2013fmeasure} to obtain the F-measure, where $\mu$ and $\epsilon$ are the mean and standard deviation within each co-saliency map respectively, and the precision and recall rates averaged over all samples are combined as defined below:
\begin{equation}
F_\beta = \frac{(1+\beta^2) \text{Precision} \times \text{Recall}}{\beta^2 \text{Precision} + \text{Recall}}
\end{equation}
where $\beta^2=0.3$ as typically used in the literature.

The deep saliency networks are implemented with the \textit{Caffe} package \cite{jia2014caffe} to follow the publicly available model of the DeepLab system. For the IrISnet, we used a large single saliency detection dataset, MSRA10K \cite{liu2007msra10k}, and the size of input images (i.e., $321 \times 321$) and hyper parameters for training are set as suggested in \cite{li2016_DCL}. The IeISnet consists of sequential fully-connected, batch normalization \cite{ioffe2015batnorm}, and rectified linear unit (ReLU) layers, which are trained with several co-saliency detection datasets, i.e., Cosal2015 \cite{zhang2016ldw} and CPD \cite{li2011co}, including either of iCoseg or MSRC that is not used for testing to exploit as much training data as possible. Even though parts of the Cosal2015 dataset (e.g., \textit{baseball}) tend to put far more emphasis on the correspondence than on the intra-image saliency, it is acceptable to our IeISnet training because the definition of IeIS also focuses more on the correspondence.
We set the learning rate and momentum parameter to 0.001 and 0.9 respectively, and the weight decay is 0.0005. As in \cite{yang2013_GMR,zhang2016ldw,li2016_DCL}, we set $n_m$ for each $I^m$ to 200, where 150 and 50 superpixels at different scales are additionally used for the IeIS detection, and the precise value of $n_m$ is determined by the SLIC algorithm. We consistently set $K=100$ and $k=5$ irrespective of the number of images $M$. For the IeISnet learning, we set $\rho=0.7$ considering the number of true positives and negatives in the training co-saliency detection datasets, and $\gamma$ is empirically set to 3. The parameter $\alpha$ is usually set to 0.99 in the literature, but we use $\alpha=0.95$ since the seed propagation steps are performed with more reliable foreground seeds, and we set $\eta=2$ for the same reason. Lastly, we use $\sigma=0.25$ and $\tau=0.5$, and also conduct grid search experiments for $\alpha$, $\eta$, $\sigma$, and $\tau$ to ensure that we select the appropriate values of these parameters.

\bgroup
\def\arraystretch{1.3}
\begin{table}[htb!]
	\centering
	\caption{Average execution time per image.
	}
	\begin{tabular}{l *{7}{c}}
		\thickhline
		 & CB & HS & MG & LDW & MIL & DIM* & Ours \\
		\hline
		
		Time (s) & 1.02 & 103.36 & 1.12 & 6.52 & 12.25 & 19.6 & 1.79 \\
		\thickhline
	\end{tabular} \\
	\justifying
	\vspace{0.1cm}
	* The running times of DIM and HS, cited from \cite{zhang2016transfer,liu2014hs}, were measured \\
	\hphantom{aaaa} only with iCoseg.
	\label{tab:runtime}
\end{table}

\subsection{Run Time Comparison}
We conduct the experiments with our unoptimized code run on a PC with Intel i7-6700 CPU, 32GB RAM, and GTX Titan X GPU. The code is implemented in MATLAB except for the SLIC algorithm in C++, and the GPU acceleration was applied only for the \textit{Caffe} framework. Table~\ref{tab:runtime} lists the average execution time per image using several different methods, where the execution times of LDW, MIL, DIM, and HS are cite from their papers. The first two values were measured using a PC with two 2.8GHz 6-core CPUs, 64GM RAM, and GTX Titan black GPU in \cite{zhang2016ldw,zhang2016mil}, the third one with Intel i3-2130 CPU and 8GB RAM in \cite{zhang2016transfer}, and the last one with Intel i7-3770 and 4GB RAM in \cite{liu2014hs}. As can be seen, the proposed method has moderate computational complexity with state-of-the-art performance as evaluated below. In particular, our method runs faster leveraging the supervised learning schemes compared to the other ones based on the online weakly supervised learning \cite{zhang2016mil,zhang2016ldw,zhang2016transfer}, and shows the execution time similar to that of the efficient CB \cite{fu2013cb} and MG \cite{li2015mg} methods.

\begin{figure*}[htb!]
	\centering
	
	\includegraphics[width=0.80\linewidth,height=0.6\linewidth]{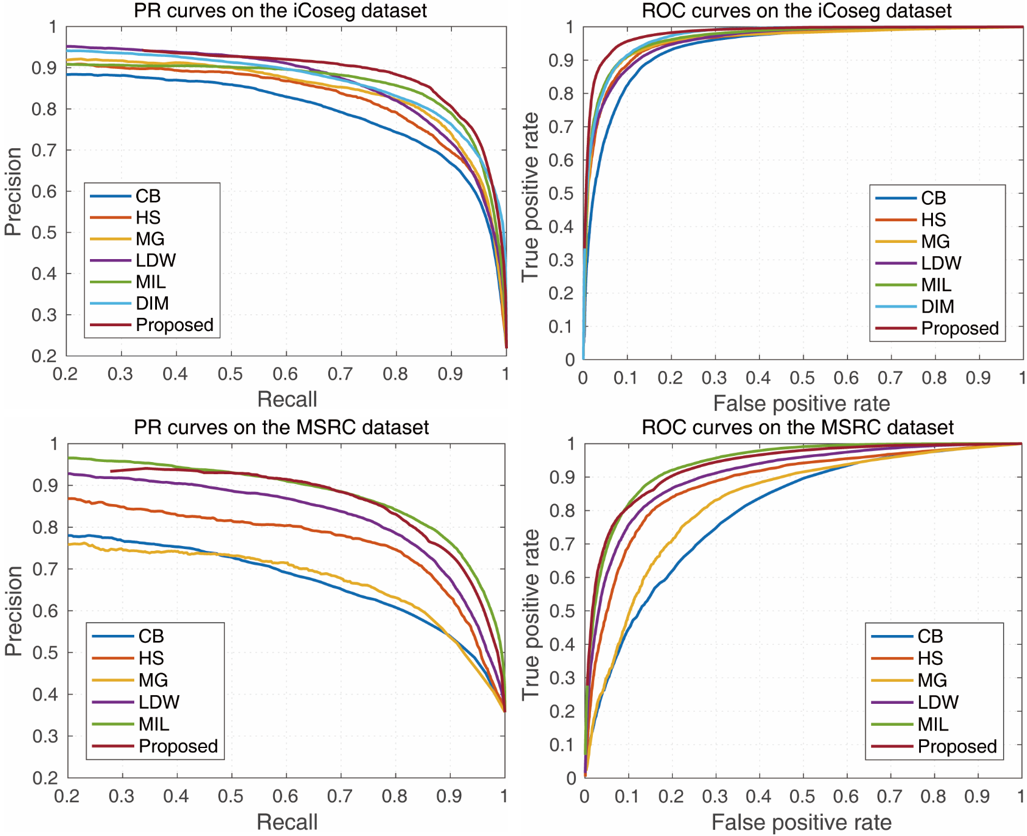}
	
	\caption{Quantitative comparison on the iCoseg and MSRC datasets with the PR and ROC curves.}
	\label{fig:quancmp}
\end{figure*}

\subsection{Comparison with the State-of-the-Art}
With the evaluation criteria stated above, we compare the proposed co-saliency detection method with other major algorithms ranging from the bottom-up ones based on handcrafted features to the learning-based ones using high-level features: CB \cite{fu2013cb}, HS \cite{liu2014hs}, MG \cite{li2015mg}, LDW \cite{zhang2016ldw}, MIL \cite{zhang2016mil}, DIM \cite{zhang2016transfer} (only for the iCoseg dataset). At first, Figs.~\ref{fig:viscmp_icoseg} and \ref{fig:viscmp_msrc} show several visual examples of resulting co-saliency maps for qualitative comparison, where it can be seen that the proposed method more uniformly detects co-salient objects and better suppresses background regions than the others do. In particular, the \textit{Alaskan bear} set shows the background regions similar to co-salient objects in color. Even though our auxiliary co-saliency maps focus on the color similarity, emphasizing the background seeds moderately suppresses the background noise that it may bring about. The MG method effectively finds the co-salient objects consistent in color, e.g., \textit{Red Sox player}, but it has weaknesses in suppressing noisy backgrounds and detecting the co-salient regions inconsistent in color, as shown in Fig.~\ref{fig:viscmp_msrc}. The \textit{Statue of liberty} set includes a lot of the co-salient regions that are not salient in terms of single-image saliency. The most representative case is the first image, where only the torch probably looks salient, but every part of the statue is co-salient in the group. Because each image in the MSRC dataset probably contains a single co-salient object, it is effective to first find salient regions in terms of single-image saliency and then analyze the correspondence in each group. Thus, the results of the proposed method show well-suppressed common backgrounds.

For the quantitative comparison, Fig.~\ref{fig:quancmp} shows the PR and ROC curves, and Table~\ref{tab:quancmp} contains the AP, AUC, and F-measure values of ours and compared methods. As for the iCoseg dataset, the proposed method outperforms the other ones on all the evaluation criteria. Both the PR and ROC curves show that our co-saliency maps result in the highest precision/recall rates in the widest ranges of recall/false positive rates, especially at decent recall/false positive rates ($\sim$ 0.8/0.1). Even though, as for the MSRC dataset, our method results in the slightly low PR and ROC curves than those of MIL, it shows the best score on the F-measure criterion. As can be seen in Figs.~\ref{fig:viscmp_icoseg} and \ref{fig:viscmp_msrc}, the proposed algorithm produces more assertive co-saliency maps than those of the other methods, which has its strengths and weaknesses. Because there is a certain amount of detected regions for each image even with a threshold close to 1, it guarantees a degree of recall, but there is also a limit to precision/false positive rates. However, it is noteworthy that this property helps to easily determine a threshold to segment a given co-saliency map for other tasks, e.g., co-segmentation. The self-adaptive thresholds used for the F-measure evaluation also probably result in robust segmentation given assertive co-saliency maps. Table~\ref{tab:quancmp} shows that the standard deviation of F-measures is the smallest when our approach is applied, which means that our method is more robust to the variation of the binarization threshold. Meanwhile, when comparing the proposed method with the other learning-based ones, LDW, MIL, and DIM, it should be noted that they need similar neighbors from other external groups rather than a target image group and perform the weakly supervised learning given the input images for testing. These procedures assume that the external groups do not include common co-salient objects with the target group so that they can give negative samples illustrating background features in the target group. Thus, as can be seen in Table~\ref{tab:runtime}, they require relatively high computational complexity, and the fact that they need the external groups may also be a limitation in conditions where the similar neighbors are lacking or insufficient. Despite the differences in requirements, we can observe that the proposed method shows better or competitive performance compared to the other learning-based algorithms.

\bgroup
\def\arraystretch{1.2}
\begin{table}[htb!]
	\centering
	\caption{Quantitative comparison on the iCoseg and MSRC datasets with AP, AUC, and F-measures.
	}
	\begin{tabular}{l l *{4}{c}}
		\thickhline
		Dataset & Method & AP & AUC & F-measure & $\sigma_F$* \\
		\thickhline
		
		\multirow{7}{*}{\shortstack{iCoseg}}
		& CB & 0.806 & 0.937 & 0.741 & 0.145 \\
		& HS & 0.839 & 0.955 & 0.755 & 0.189 \\
		& MG & 0.854 & 0.957 & 0.794 & 0.114 \\
		& LDW & 0.875 & 0.957 & 0.799 & 0.168 \\
		& MIL & 0.866 & 0.965 & 0.814 & 0.141 \\
		& DIM & 0.877 & 0.969 & 0.792 & 0.212 \\
		& Ours & \textbf{0.896} & \textbf{0.979} & \textbf{0.823} & \textbf{0.077} \\
		\hline
		
		\multirow{6}{*}{\shortstack{MSRC}}
		& CB & 0.689 & 0.798 & 0.577 & 0.170 \\
		& HS & 0.785 & 0.882 & 0.709 & 0.197 \\
		& MG & 0.688 & 0.827 & 0.635 & 0.133 \\
		& LDW & 0.842 & 0.908 & 0.767 & 0.178 \\
		& MIL & \textbf{0.894} & \textbf{0.940} & 0.796 & 0.138 \\
		& Ours & 0.876 & 0.934 & \textbf{0.811} & \textbf{0.054} \\
		
		\thickhline
	\end{tabular}
	\\ \justifying \vspace{0.1cm}
	* $\sigma_F$ denotes a standard deviation of F-measures with the variation of the \\
	\hphantom{aaaa} binarization threshold.
	\label{tab:quancmp}
\end{table}

\begin{figure*}[htb!]
	\centering
	\includegraphics[width=\linewidth]{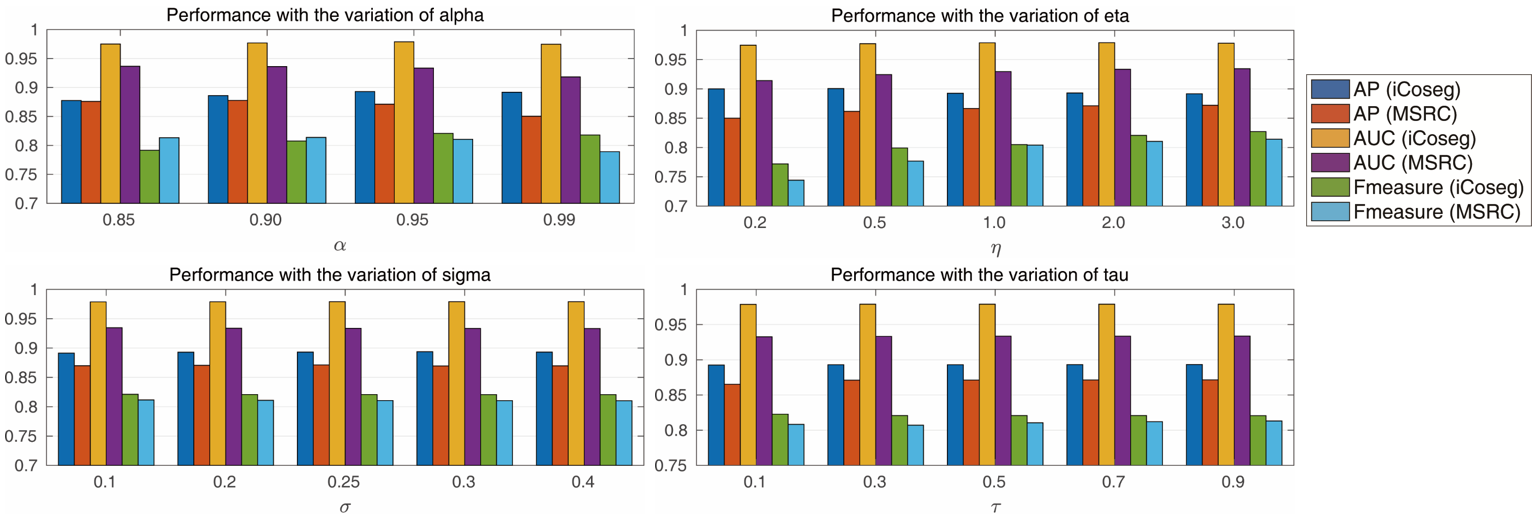}
	\caption{Grid search analysis on the parameters $\alpha$, $\eta$, $\sigma$, and $\tau$. The performance with the variations of $\alpha$ and $\eta$ slightly depends on target datasets (i.e., iCoseg and MSRC), while the proposed system is not sensitive to $\sigma$ and $\tau$.}
	\label{fig:params}
\end{figure*}

\subsection{Parameter Analysis}
We conduct the grid search experiments to find the appropriate values of $\alpha$, $\eta$, $\sigma$, and $\tau$. Fig.~\ref{fig:params} shows the AP, AUC, F-measure scores along with certain ranges of these parameters. As can been seen, the most effective value of $\alpha$ is lower with MSRC than with iCoseg, which is related to the tendency that correspondence cues are of lower importance so the fitting constraint is more emphasized for the seed propagation over the integrated graph in MSRC than in iCoseg. Likewise, the variation of the parameter $\eta$ also slightly influences the performance depending on the target datasets. Because the foreground and background seeds are basically of equal importance, one could set $\eta$ to 1, but the value of $\eta$ larger than 1 is more effective with reliable foreground seeds, especially with respect to the F-measures. On the other hand, the proposed method is not sensitive to the parameters $\sigma$ and $\tau$, so we can select the decent values for these parameters and obtain the stable results with them. Even though, in (\ref{eq:initcosal}), two different operations are applied according to $\tau$, both sides would emphasize the IeIS with large difference between IrIS and IeIS values; when they are similar to each other, they both would give almost equal contributions to the resulting initial co-saliency value. Thus, slight variations of the parameter $\tau$ do not bring about big differences in the performance of our method. The control parameter $\sigma$ for the construction of the integrated graph behaves similarly to $\tau$, where larger $\sigma$ more facilitates the seed propagation between the cluster centers and intra-image nodes with similar colors, and vice versa. It is because the various colors of co-salient objects could be reflected in the cluster layer, where different inter-image nodes represent diverse colors, with sufficiently large $K$, and the regions within an image group similar in color could share their co-saliency information through the seed propagation. 



\subsection{Co-segmentation Experiments}
Co-segmentation is a direct higher-level application of the co-salient object detection, where it can replace user interaction and provide useful prior knowledge of target objects. For example, Quan et al. \cite{quan2016gofmr} proposed to construct a graph including input images and generate two types of probability maps using low- and high-level features through graph-based optimization. For each image, the resulting two probability maps are combined by multiplication, and then a graph cut approach produces the final co-segmentation results. These probability maps could be replaced with co-saliency maps and in fact, Fu et al. \cite{fu2013cb} applied their co-saliency detection method to co-segmentation through Markov random field optimization. The approach of Chen et al. \cite{chen2014enrich} groups input images into aligned homogeneous clusters and then merges them into visual subcategories, where a discriminative detector for each subcategory is trained to find target objects within a test dataset. For each cluster, a co-segmentation method is applied to segment out the aligned objects, and this step could also be performed with co-saliency detection.

Thus, we conduct co-segmentation experiments to compare our results with those of other approaches. Two datasets, Internet-100 \cite{rubinstein2013internet} and iCoseg, are used for the evaluation with the Jaccard index (J, intersection-over-union for the foreground regions) and Precision (P, the proportion of correctly labeled pixels). We convert our co-saliency maps into co-segmentation results by simply thresholding with 0.5. Because, as for the Internet-100 dataset, there are several noisy images that do not contain target objects in each class, we normalize each auxiliary co-saliency map by the operation $x \rightarrow (x+1)/2 $ instead of normalizing it to $[0,1]$, which forces the maximum in it to be $1$. Table~\ref{tab:coseg} and Fig.~\ref{fig:coseg} show the quantitative comparison and several visual examples of our results on the Internet-100 dataset, respectively. The proposed method with simple thresholding outperforms other state-of-the-art co-segmentation algorithms or produces competitive results compared to them. Fig.~\ref{fig:coseg} shows several quality co-segmentation results, but the noisy objects have not been perfectly suppressed in the last images of the first and second rows.

\bgroup
\def\arraystretch{1.2}
\begin{table}[htb!]
	\centering
	\caption{Quantitative comparison of co-segmentation methods.
	}
	\begin{tabular}{c *{6}{c}}
		\thickhline
		\multirow{2}{*}{\shortstack{Internet-100}} & \multicolumn{2}{c}{Airplane} & \multicolumn{2}{c}{Car} & \multicolumn{2}{c}{Horse} \\
		\cline{2-7}
		& P (\%) & J (\%) & P (\%) & J (\%) & P (\%) & J (\%) \\
		\thickhline
		
		\cite{joulin2012coseg} & 47.5 & 11.7 & 59.2 & 35.2 & 64.2 & 29.5 \\
		\cite{rubinstein2013internet} & 88.0 & 55.8 & 85.3 & 64.4 & 82.8 & 51.3 \\
		\cite{chen2014enrich} & 90.3 & 40.3 & 87.7 & 64.9 & 86.2 & 33.4 \\
		\cite{quan2016gofmr} & 91.0 & 56.3 & 88.5 & 66.8 & 89.3 & 58.1 \\
		Ours & \textbf{92.5} & \textbf{58.5} & \textbf{91.0} & \textbf{74.9} & \textbf{89.6} & \textbf{59.3} \\
		
		\thickhline
	\end{tabular}
	\\ \vspace{0.1cm}
	\begin{tabular}{l *{4}{c}}
	\thickhline
	iCoseg & \cite{keuttel2012coseg} & \cite{faktor2013coseg} & \cite{quan2016gofmr} & Ours \\
	\thickhline
	P (\%) & 91.4 & 92.8 & 93.3 & \textbf{93.9} \\
	J & -- & 0.73 & \textbf{0.76} & 0.75 \\
	
	\thickhline
	\end{tabular}
	\label{tab:coseg}
\end{table}

\begin{figure}[htb!]
	\centering
	\includegraphics[width=\linewidth]{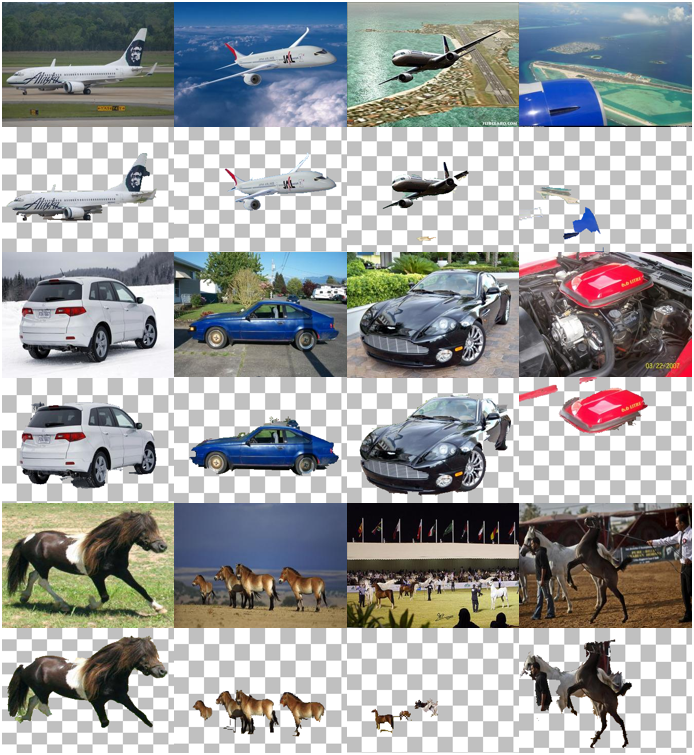}
	\caption{Visual examples of our co-segmentation results on the Internet-100 dataset. Top to bottom: Airplane, Car, Horse classes.}
	\label{fig:coseg}
\end{figure}

\section{Conclusions}
We have proposed a co-saliency detection method, which finds the regions of high initial co-saliency values with the deep saliency networks and complementary regions through the seed propagation over the integrated graph. Given salient regions within each image in terms of single-image saliency, the features extracted from these foregrounds in a group are concatenated with the descriptor of each segment to be fed into the inter-image saliency network. The resulting IrIS and IeIS values are combined to produce initial co-saliency maps, which then provide foreground and background seeds for the seed propagation steps. The unified graph is constructed with the affinity matrices using color similarities, and the newly propagated co-saliency values become complementary components for the final co-saliency maps. The experimental results indicate that the proposed method shows the state-of-the-art performance with decent requirements about computational complexity and input images/groups.


%

%

%
%

\ifCLASSOPTIONcaptionsoff
  \newpage
\fi



\bibliographystyle{./IEEEtranBST/IEEEtran.bst}
\bibliography{IEEEabrv.bib,cosal_refs.bib}
\end{document}